\pdfoutput=1 

\newif\ifsubmit
\submittrue 

\documentclass[11pt]{article}

\usepackage[preprint]{acl}

\usepackage{times}
\usepackage{latexsym}
\usepackage[T2A,T1]{fontenc} 

\usepackage[utf8]{inputenc} 
\usepackage[ukrainian,english]{babel}
 
\usepackage{microtype} 

\usepackage{inconsolata} 
\usepackage{csquotes}
\usepackage{booktabs} 
\usepackage{tabularx}
\usepackage{graphicx} 
\usepackage{comment}
\ifsubmit\else 
\fi

\title{Negation: A Pink Elephant in the Large Language Models' Room?}

\author{Tereza Vrabcová \and Marek Kadlčík \and Petr Sojka \and Michal Štefánik \and Michal Spiegel \\
         Faculty of Informatics, Masaryk University}

\providecommand{\allowhyphens}{\ifvmode\else\nobreak\hskip0pt\relax\fi}
\newcommand\TT[1]{\multicolumn{1}{l}{\textsf{\small\begin{tabular}{@{}l@{}}#1\end{tabular}}}}
\def\B#1\%{\textbf{#1}\%}
\def\M#1&{{\footnotesize\sf#1}&}
\newcommand{\LM}[1]{\llap{\ensuremath{#1}}}
\bgroup 
  \catcode`\_=\active
  \gdef\works{\catcode`_=\active
      \def_{\setbox0=\hbox{0}\hspace*{\wd0}\relax}}
\egroup
\newcommand{\nstress}[1]{\textbf{\textit{#1}}} 
\let\stress\emph 
\ifsubmit \newcommand{\todo}[1]{}
\else \newcommand{\todo}[1]{\textcolor{red}{TODO: #1}\PackageWarning{TODO:}{#1!}}
\fi

\begin{document}
\maketitle
\begin{abstract}  
Negations are key to determining sentence meaning, making them essential for logical reasoning. 
Despite their importance, negations pose a substantial challenge for large language models (LLMs) and remain underexplored.

We constructed and published two new textual entailment datasets NoFEVER-ML and NoSNLI-ML in four languages (English, Czech, German, and Ukrainian) with \textit{paired} examples differing in negation.
It allows investigation of the root causes of the negation problem and its exemplification: how popular LLM model properties and language impact their inability to handle negation correctly.

\smallskip

Contrary to previous work, we show that increasing the model size may improve the models' ability to handle negations. 
Furthermore, we find that both the models' reasoning accuracy and robustness to negation are language-dependent and that the length and explicitness of the premise have an impact on robustness.

There is better accuracy in projective language with fixed order, such as English, than in non-projective ones, such as German or Czech.

Our entailment datasets pave the way to further research for explanation and exemplification of the negation problem, minimization of LLM hallucinations, and improvement of LLM reasoning in multilingual settings.
\end{abstract}

\section{Introduction}

Negation, a cornerstone of logical reasoning and nuanced communication, often remains a `pink elephant in the room' for Large Language Models (LLMs).
Despite their impressive advancements, effectively understanding and processing negation poses a fundamental, and often underestimated, challenge to their reliability and logical reasoning capabilities. 
Processing negation is critical for reliable performance, and leads to hallucinations as \citet{neg:varshney-etal-2025-investigating} showed on four tasks with negation: `false premise completion', `constrained fact generation', `multiple choice question answering', and `fact generation'.
Vision-language models do not understand negation either~\cite{neg:alhamoud2025visionlanguagemodelsunderstandnegation}: when asked to provide a picture with no elephant in the room, they put an elephant there anyway.
It appears that negation tokens (e.g., \enquote{not}) have a limited effect on the representations learned distributionally~\cite{neg:singh2024learnnosayyes}.

This difficulty, sometimes termed \nstress{negation blindness} or \nstress{NO syndrome}, is well-documented. Studies show LLMs may struggle to distinguish between facts and their negations, misunderstand the semantic impact of negative particles, and fail to generalize negation handling robustly, even with instruction tuning.

\begin{figure}[t]
    \centering
    \includegraphics[width=1.0\linewidth]{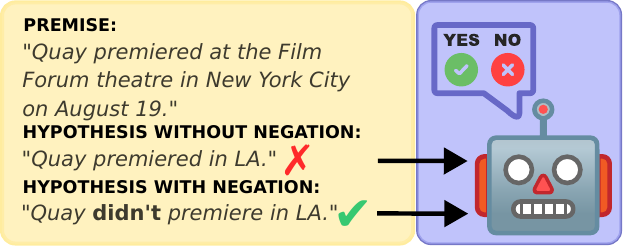}
    \caption{We extend existing entailment datasets FEVER and SNLI with negated hypotheses in four languages (ENG, UKR, CZE, GER) and measure how negations degrade the accuracy of model reasoning.}
    \label{fig:vis-abstract}
\end{figure}

\subsubsection*{Prior Work on Negation in LLMs}
The challenge of negation for LLMs has been approached from several angles, as in the parable about the elephant met by blind monks.
We categorize prior work into the following areas:

\paragraph*{1. Foundational Issues \& Early Observations}
Early research highlighted fundamental difficulties for language models in capturing logical operations such as negation within continuous vector space representations \cite{nlp:hermann-etal-2013-bad}. 
Subsequent work observed that LLMs can be equally prone to generating factual statements and their incorrect negations~\cite{neg:kassner-schutze-2020-negated-probes}, and specific assessments were made on models like BERT regarding their sensitivity to the meaning conveyed by negation~\cite{neg:ettinger-2020-what-bert-is-not}.

\paragraph*{2. Datasets \& Benchmarks for Negation}
Several benchmarks have been developed to probe negation understanding. For instance, the NeQA dataset was introduced to test question answering with negation, where models did not exhibit straightforward positive scaling~\cite{neg:zhang-etal-2023-beyond}. Another large-scale benchmark was created to challenge LLMs with a wide array of negative sentences, finding that while fine-tuning helps, generalization in handling negation remains elusive~\cite{neg:garcia-ferrero-etal-2023-dataset}. The NaN-NLI test suite specifically focuses on sub-clausal negation~\cite{neg:truong-2023-naysayers}. \citet{neg:mondorf-plank-2024-liar} introduced a benchmark for suppositional reasoning that convincingly shows that ``proficient models struggle with inferring the logical implications of potentially false statements''.

\paragraph*{3. Analysis of LLM Failures \& Characteristics of Negation Handling}
Researchers have investigated whether the semantic constraints of function words like negation are adequately learned and how context impacts their embeddings~\cite{neg:kalouli-etal-2022-negation}. It has been shown that LLMs tend to underestimate the significant impact of negation on sentence meaning~\cite{neg:anschutz-etal-2023-correct}. A broad analysis across various LLMs, including instruction-tuned models, confirmed their inability to capture the lexical semantics of negation and to reason effectively under negation, with model size not initially appearing as a clear mitigating factor~\cite{neg:truong-2023-naysayers} (this paper re-evaluates the model size aspect). Furthermore, negation tokens like ``not'' appear to have a limited effect on the distributional representations learned by models~\cite{neg:singh2024learnnosayyes}. The phenomenon often leads to hallucinations in tasks involving negation, such as false premise completion and constrained fact generation \cite{neg:varshney-etal-2025-investigating}. Other studies have also investigated the limits of LLMs on general compositional tasks, which implicitly include negation \cite{nlp:dziri-2023-faith-fate}.

\paragraph*{4. Negation in Multimodal \& Vision-Language Models}
The challenge of negation extends beyond text-only models. Vision-language models have also been shown to struggle with understanding negation, for example, when prompted to generate images \stress{without} a certain object \cite{neg:alhamoud2025visionlanguagemodelsunderstandnegation}. Efforts are being made to address this, such as proposing negation context-aware reinforcement learning feedback loops to ensure generated text and images correctly reflect negated queries~\cite{neg:nadeem2024negationblindnesslargelanguage}.

\paragraph*{5. Mitigation Strategies \& Improving Negation Handling}
Various strategies have been proposed to improve negation handling. These include augmenting training data with more negative examples~\cite{helwe-etal-2022-tina}, paraphrasing input text into affirmative terms to remove explicit negation while preserving meaning \cite{rezaei-blanco-2024-paraphrasing}, and exploring fine-tuning processes, including instruction prompt adjustments, to explicitly account for negation \cite{nli:anschutz-etal-2023-correct, neg:truong-2023-naysayers}.
\citet{nlp:castricato-2024-suppressing-pink-elephants-direct} study controllable generation of LLM through the lens of the ``Pink elephant paradox''~\cite{psy:spiers-2002-pink-elephant-paradox} and think that form of Reinforcement Learning from AI Feedback (RLAIF) that they term \stress{Direct Principle Feedback} assesses the problem. 

\paragraph*{6. Cross-lingual \& Multilingual Negation Studies}
The study of negation awareness has also been extended to multilingual contexts. For example, \citet{neg:hartmann-etal-2021-multilingual} developed a multilingual benchmark to probe negation-awareness across languages including English, Bulgarian, German, French, and Chinese, setting a precedent for cross-lingual investigations in this domain.
\citet{neg:vrabcova-sojka-2024-negation-czech} develop CsFEVER dataset and confirm that negation causes problems in Czech as well.

\medskip 

In this paper, we focus on the effects of moderating variables on robustness to negation, posed by the following research questions:
\begin{description}
\item[RQ1:] Can the \stress{scale of model size} improve models’ ability to handle negations? 
\newline
In one of the findings of~\citet{neg:truong-2023-naysayers}, it is posited that the LLMs are unable to reason under negation, with the increased size of the LLM not being a mitigating factor. We review this claim by evaluating models from the same LLM family with different model sizes.

\item[RQ2:] Is the models' capability to handle negations \stress{language-dependent}?
\newline
As most studies concerning negation in LLMs focus on English datasets, we extend our scope to include multiple languages with differing levels of word-order flexibility.
\end{description}

The primary contributions of this work are:
\begin{itemize}
\item The development and release of two novel, multilingual textual entailment datasets specifically designed to evaluate negation. These datasets span English, Czech, German, and Ukrainian and feature paired examples differing only in negation.

\item Empirical evidence demonstrates that larger model sizes can indeed improve the handling of negations, offering a counterpoint to some earlier research.
An analysis reveals that while linguistic features of a language play a role, the specificity and length of the premise context have a more pronounced impact on an LLM's robustness to negation than the language itself.
\end{itemize}

The paper is structured as follows. 
Section~\ref{sec:dataset} presents the preparation of our two new NLI datasets.
Section~\ref{sec:methodology} describes the methodology and the LLMs we use in our experiments.
Results are discussed in Section~\ref{sec:results}.
We conclude and outline future directions in Section~\ref{sec:conclusions}.

\section{Datasets}
\label{sec:dataset}

We evaluate the accuracy of the models in our experiments using two datasets that have been originally formatted for the Natural Language Inference (NLI) task, i.e., containing premise-hypothesis pairs and the label denoting their logical relationship (entailment, neutral, contradiction).

\begin{description}
\item[SNLI]\cite{nlp:bowman-etal-2015-snli} is a popular NLI dataset focusing on the evaluation of general and common sense knowledge of the LLMs.
The premises are captions of images capturing everyday situations and, as such, are quite short, consisting of one to two sentences.
The hypotheses have been generated manually by human annotators.

\item[FEVER-NLI]\cite{nli:thorne-etal-2018-fever} is a modification of the FEVER dataset, which originally focused on the task of fact extraction and verification.
The dataset consists primarily of factual statements retrieved from Wikipedia.
\end{description}

As our aim is to study the effects of negation on the accuracy of the models, we modify the format of these datasets for the purpose of the TE task and extend them to contain the negation of the original hypothesis.
The original datasets contain a low percentage of hypotheses with common negation markers, such as \textit{no, n't, not, nobody, neither,} or \textit{nothing}, and differ in the verbosity of premises, as seen in the Table~\ref{tab:SNLIandFEVER-NLIdata}. 

In Table~\ref{tab:preparedNLIdata}, we include average lengths of premises and hypotheses of the newly prepared datasets for comparison, NoFEVER-ML and NoSNLI-ML.

\begin{table}[ht]
\tabcolsep3dd
\begin{tabular}{@{}rrr@{}}
\toprule
\multicolumn{1}{r}{} &
  \TT{SNLI} &
  \TT{FEVER-NLI} \\
\midrule
Rows in the dataset    & 20,000  & 10,000\\
Rows w/ negation       &    868  & 189\\ 
\% of rows w/ negation &   4.34  & 1.89\\
Premise word count     &  13.94  & 51.96\\
Hypothesis word count  & \hphantom{0}7.50 &  \hphantom{0}8.80 \\
\bottomrule
\end{tabular}
\caption{Comparison of the original English SNLI and FEVER-NLI datasets}
\label{tab:SNLIandFEVER-NLIdata}
\end{table}

\begin{table*}
\tabcolsep2dd
\begin{tabular}{@{}rrrrrrrrrr@{}}
\toprule
\multicolumn{1}{r}{} &
  \TT{CES\\FEVER\\MANUAL} &
  \TT{CES\\FEVER} &
  \TT{CES\\SNLI} &
  \TT{ENG\\FEVER} &
  \TT{ENG\\SNLI} &
  \TT{DEU\\FEVER} &
  \TT{DEU\\SNLI} &
  \TT{UKR\\FEVER} &
  \TT{UKR\\SNLI} \\
  \midrule

Premise (chars) & 268.36 & 265.06 & 51.71 & 266.31 & 58.02 & 306.60 & 70.33 & 278.54 & 58.55 \\
Hypothesis without negation (chars) & 42.87 & 41.69 & 24.06 & 42.12 & 28.71 & 49.63 & 34.55 & 43.38 & 27.71 \\
Hypothesis with negation (chars) & 44.89 & 43.58 & 26.42 & 51.13 & 37.26 & 53.69 & 38.87 & 45.11 & 29.70 \\
Premise (words) & 49.36 & 48.85 & 10.74 & 54.12 & 14.05 & 55.43 & 13.58 & 47.90 & 10.55 \\
Hypothesis without negation (words) & 7.86 & 7.66 & 5.20 & 8.72 & 7.13 & 8.97 & 7.01 & 7.35 & 4.99 \\
Hypothesis with negation (words) & 7.86 & 7.69 & 5.32 & 10.95 & 9.23 & 9.82 & 7.93 & 8.34 & 5.99 \\
\bottomrule
\end{tabular}
\caption{Comparison of average lengths across evaluated datasets, NoFEVER-ML and NoSNLI-ML.}
\label{tab:preparedNLIdata}
\end{table*}

\subsection{Preparation of new datasets in English}

In order to prepare new datasets containing negation, we first sanitize the datasets and then extend the datasets with hypotheses including negation. Contrary to \cite{neg:gururangan-etal-2018-annotation}, we do not distinguish between contradiction and logical negation of tokens expressing negation.
We then manually replace incorrectly generated sentences and create 2~datasets. 

\begin{description}
    \item[NoFEVER-ENG dataset:] This dataset contains \mbox{2,600\,premise-hypothesis-negated\_hypothesis} triplets, with 1,513 premises entailing the hypothesis without negation, and 1,087 premises entailing the hypothesis with negation.
    \item[NoSNLI-ENG dataset:] This dataset contains 6,261 triplets, with 3,245 premises entailing the hypothesis without negation, and 3,016 premises entailing the hypothesis with negation.
\end{description}

\paragraph*{Sanitization and Filtration}

Firstly, we sanitize the datasets by removing all incomplete or invalid rows, i.e., removing rows with empty premises, empty hypotheses, or invalid labels. Secondly, we remove rows that have the \textit{neutral} entailment label, as the effect of negation on their entailment value is not uniform.

\paragraph*{Negation}

We employ the \texttt{negate} Python module for the first phase of our negation process. This module is a rule-based negation tool for the English language, utilizing the \texttt{spaCy} model \texttt{en\_core\_web\_md} for POS tagging and dependency parsing in order to correctly locate the verb in the sentence.
The verb is then negated by the inclusion of an auxiliary verb and a negation marker, with our preference in the module set to the contraction \textit{n't}. 
The tool is able to negate in both directions, meaning that it can remove negation from the verb as well.

Rows with hypotheses that do not contain verbs are discarded during the generation process.

\paragraph*{Manual verification}

As the tool we used for the hypothesis generation is rule-based, it comes with certain limitations and cases that are not covered by the rules and have to be fixed manually. For the second phase of our negation process, we have performed a full manual check to ensure the correctness of the newly created hypotheses.
\smallskip\noindent

\textbf{\textit{Some, any}, nor \textit{yet} are properly supported.}\\
Example 1:\\
\textit{Input:} There is someone jumping. \\
\textit{Negate output:} There isn't \underline{someone} jumping. \\
\textit{Manual change:} There isn't \underline{anyone} jumping.

\noindent Example 2:\newline
\textit{Input:} Keegan-Michael Key has yet to play anyone.\\
\textit{Negate output:} Keegan-Michael Key \underline{doesn't have} yet to play anyone.\\
\textit{Manual change:} Keegan-Michale Key \underline{has played} \underline{someone}.
\smallskip\noindent

\textbf{The rules of the module as aimed at the \textit{Not-negation}, most likely due to its higher frequency in the English language, in comparison to the \textit{No-negation}.}

\noindent Example 3:\newline
\textit{Input:} There is no one running. \\
\textit{Negate output:} There \underline{isn't} no one running. \\
\textit{Manual change:} There is \underline{someone} running.

The typos in the original datasets have been fixed only in cases where the typo directly impacted the negation of the verb.

\subsection{Translation of new datasets into Czech, German, and Ukrainian}

To explore the impact of negation on the accuracy of the LLMs across languages as started by \cite{neg:hartmann-etal-2021-multilingual} for English, Bulgarian, German, French and Chinese, we translate our English (ENG) datasets into Czech (CES), German (DEU), and Ukrainian (UKR).
We have chosen these languages due to their differing features, such as morphology, orthography, alphabet, and projectivity, as specified in Table~\ref{tab:adata-morphology}.

\begin{table*}[ht]
\tabcolsep3.9dd
\begin{tabular}{@{}rllll@{}}
\toprule
\multicolumn{1}{r}{} &
  \TT{ENG} &
  \TT{CES} &
  \TT{DEU} &
  \TT{UKR} \\
\midrule
Language Family        & West Germanic & West Slavic  & West Germanic                      & East Slavic  \\
Script                 & Latin         & Latin        & Latin                              & Cyrillic     \\ 
Linguistic Typology    & Analytic      & Fusional     & Fusional w/ Agglutinative Features & Fusional     \\ 
Word Order Flexibility & Low           & High         & Moderate                           & High         \\
\bottomrule
\end{tabular}
\caption{Comparison of linguistic features between languages English (ENG), Czech (CES), German (DEU), and Ukrainian (UKR).}
\label{tab:adata-morphology}
\end{table*}

\paragraph{Translator Selection}

There are many machine translation tools to choose from.
To choose between Google Translate and DeepL, we have translated 100 rows from the modified English FEVER-NLI dataset into Czech, comparing the quality of translation between the translations and to a preexisting translation of the FEVER-NLI dataset that has been manually verified by a human annotator. 
From these 100 rows, DeepL provided higher quality for 44~rows, Google Translate for~30, with 26~rows having equal translation quality between them.

We carry out the Statistical Sign Test with the ties equally split and with the alpha level $\alpha = 0.05$. The null hypothesis is that both translators are of equal quality. Our result value is $p$-value = $0.2$ and thus we cannot reject our null hypothesis. 

As the difference between the translations is not statistically significant in the 100~examples, we have chosen to use DeepL to translate the datasets due to the higher percentage of winning translations.

\paragraph{Translation validation}
After we had translated the datasets into our target languages, we randomly selected 600 rows from each dataset and checked the quality of the translation. 

Although the translation quality of the model we have used is generally high, it is still an LLM that is not immune to hallucinations on rarely seen data. We have, for example, in the Ukrainian SNLI dataset, observed the outlier case shown in Table~\ref{tab:adata-outlier}.
This behavior does not occur when the sentence does not include negation.
We suppose that the combination of the inclusion of negation, the tense in the original English sentence, and the inclusion of an initialism (\textit{pb and j}) created confusion in the model. However, as the training data of the model isn't public, we cannot confirm our hypothesis.

\paragraph{Dataset Availability}
Our final datasets are publicly available in an anonymized repository~\cite{data:datasets}.

\begin{table*}[ht]
\begin{tabular}{@{}rll@{}}
\toprule
\multicolumn{1}{r}{} &
  \TT{Original/Translated Sentence} &
  \TT{Sentence in English} \\
  \midrule
ENG  & A man isn't eating pb and j              & A man isn't eating pb and j                 \\
CES  & Muž nejí pb a j.                         & A man isn't eating pb and j.                \\
DEU  & Ein Mann isst kein Pb und J              & A man isn't eating Pb and J                 \\
UKR  & \foreignlanguage{ukrainian}{Чоловік не їсть пломбір з соєвими бобами} & A man isn't eating ice cream with soy beans \\
\bottomrule
\end{tabular}
\caption{Translation of an outlier example sentence using DeepL}
\label{tab:adata-outlier}
\end{table*}

\section{Methodology}
\label{sec:methodology}

To study the effect of negation on the reasoning reliability of large language models, we evaluate them on \textit{textual entailment}~\cite{nli:Putra-2024-textual-entailment-overview}. In this task, the model receives two claims, a \stress{premise} and a \stress{hypothesis}, and decides whether the hypothesis follows from the premise.

In our experiments, we inspect whether the presence of negation in the hypothesis \stress{causes} degradation in the accuracy of the language model reasoning. For this purpose, we prepare a set of paired evaluation sentences that differ only in a negation of the primary verb. We call the variants \textbf{hypotheses without negation} and \textbf{hypotheses with negation} based on whether they contain a negation. Necessarily, if one follows from the premise, the other contradicts it, and vice versa. If the models' reasoning is robust to negations, its accuracy should be the same between the two groups.

The evaluated models are LLama~3~\cite{nlp:Dubey-etal-2024-llama3-herd-of-models} with 3B, 8B, and 70B parameters~\cite{data:llama3-4negation}, Qwen~2.5~\cite{nlp:yang-etal-2024-qwen2-technicalreport} with 1.5B, 7B, and 72B parameters~\cite{data:qwen2.5}, and Mistral Nemo 12B, Small 22B, and Large 123B~\cite{data:mistral-hf-4negation}.

All of the evaluated models are publicly available state-of-the-art models based on the \texttt{transformer} architecture~\cite{llm:vaswani2017attention}, have open weights, a multi-lingual pre-training, and are fine-tuned to follow user instructions. In the evaluation, we prompt the model to judge if a hypothesis follows from a premise or not. We let the model finish a response \enquote{The answer is:} with either \textit{True} or \textit{False} using a greedy decoding. To ensure the response follows the predefined structure, we employ Outlines~\cite{nlp:willard2023efficient} and LMFE~\cite{lm_format_enforcer}. More details of the prompt format can be found in~Appendix~\ref{appendix:prompting}.

\section{Results}
\label{sec:results}
We summarize our results in this section, including the results of the statistical tests to support these claims. We find that accuracy and negation sensitivity improve with increasing model size and the information specificity of dataset premises, while variations in languages are not statistically significant.

Unless otherwise stated, for the statistical tests, we are using an alpha level (significance threshold) $\alpha = 0.05$. More details of the specific values used during computation can be found in~Appendix~\ref{appendix:stats}.

\begin{figure*}[p]
\centerline{\includegraphics[width=.8\textwidth]{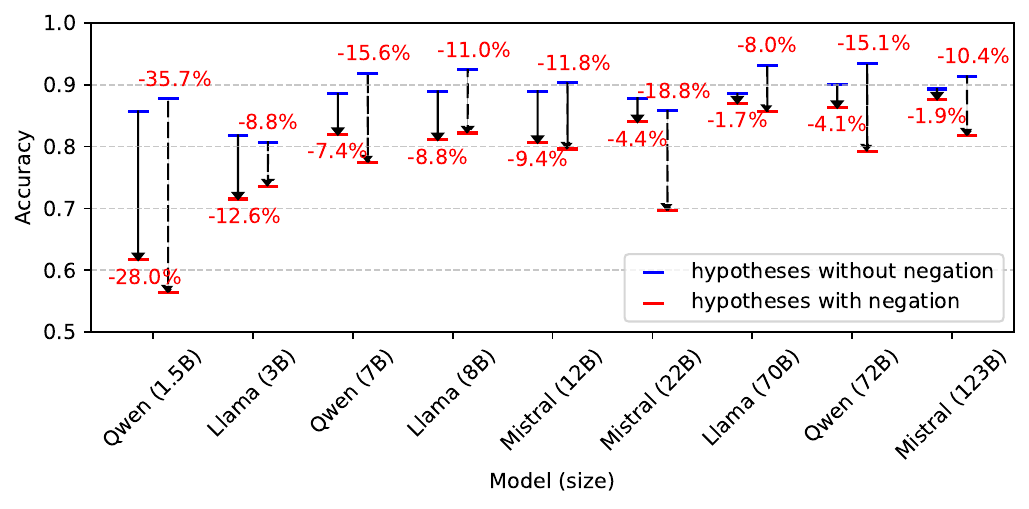}}
\vspace*{-.7\baselineskip} 
\caption{\textbf{Robustness to negations across model sizes:} Models' accuracy difference on textual entailment when facing a hypothesis (i)~including and (ii)~not including a negation. The results are an aggregate across four languages (CES, DEU, ENG, UKR) measured for FEVER (solid line; left) and SNLI (dashed line; right) datasets.}
\label{fig:permodel}
\end{figure*}

\begin{figure*}
\centerline{\includegraphics[width=0.85\textwidth]{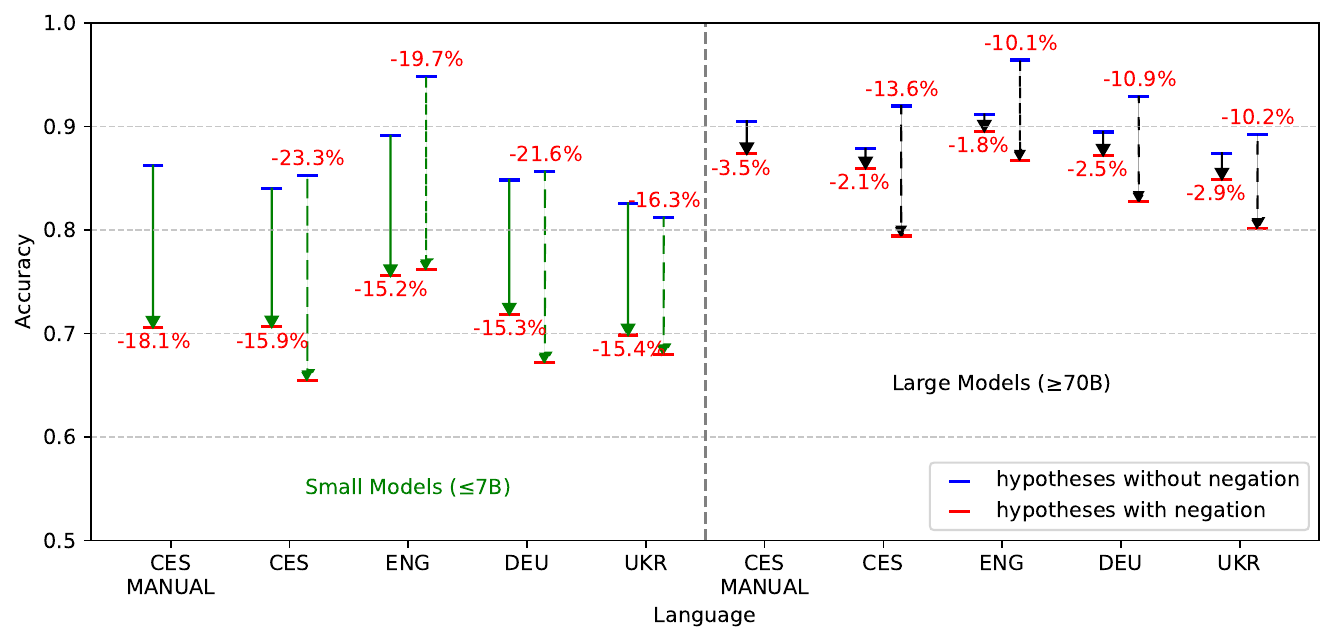}}
\vspace*{-.7\baselineskip} 
\caption{\textbf{Robustness to negations across languages:} Accuracy difference of \textit{smaller} ($\leq$7B; left) and \textit{large} models ($\geq$70B; right) on textual entailment when facing a hypothesis (i)~including and (ii)~not including a negation. Inputs are \textit{identical} across \textit{different} languages. The results are aggregated across three models in each size group, measured for two datasets: FEVER (solid line) and SNLI (dashed line).}
\label{fig:performancechange}
\end{figure*}

\begin{figure*}
\centerline{\includegraphics[width=\textwidth]{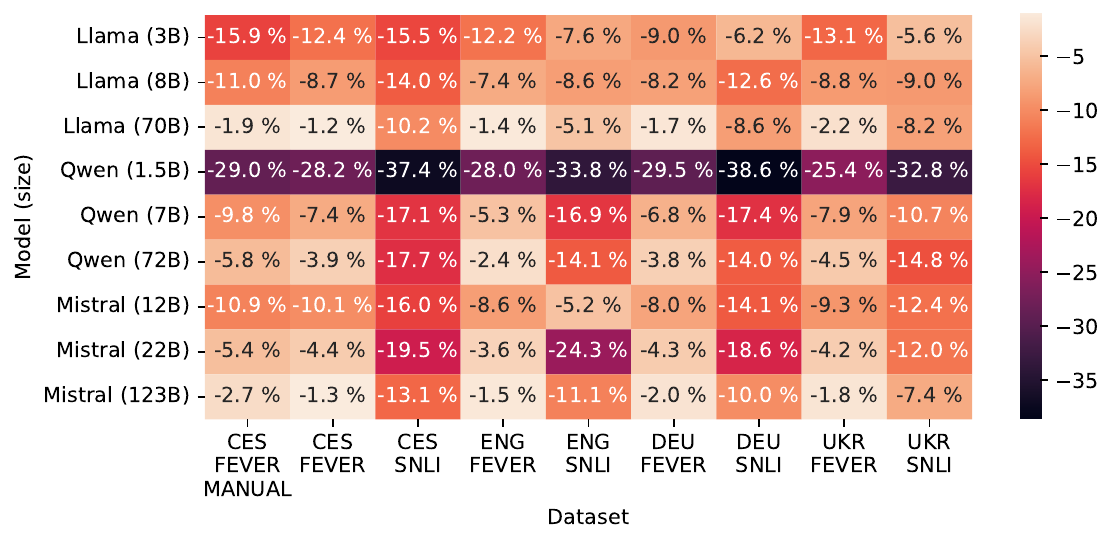}}
\vspace*{-\baselineskip}
\caption{\textbf{Robustness to negations in detail:} Accuracy difference caused by negation for all 9~evaluated models across all 4~evaluated languages and our two new datasets. The bigger models consistently handle negation better.}
\vspace*{-.7\baselineskip}
\label{fig:fullresults}
\end{figure*}

\paragraph{\textbf{Model size has a positive correlation with models' ability to handle negation.}} 

Figure~\ref{fig:permodel} shows the accuracy change of different models for the hypotheses with and without negation. We found that all evaluated models perform better than a random guess and that the larger models \textit{narrow} the gap between the accuracies, with two outliers on the SNLI dataset: a slight increase on the middle-sized Llama model (8B) and a marked drop on the middle-sized Mistral model~(22B).

Using the results shown in Figure~\ref{fig:fullresults}, we compute the correlation of the model size and the relative accuracy change for each evaluated dataset, as well as the overall correlation. Using the Shapiro-Wilk test, we find that the majority of datasets do not have a normal distribution over the different model sizes. Thus, we compute correlation using the Spearman correlation coefficient, as the Pearson correlation coefficient presumes a normal distribution of values.

The results indicate a strong positive correlation between model size and the relative accuracy change on the FEVER dataset and a weak-to-moderate positive correlation on the SNLI dataset. For an averaged relative accuracy change for all languages per model, the results indicate a strong positive correlation with the Spearman correlation coefficient of~$0.867$, supporting out RQ1.

We posit that the differences in correlation are due to different philosophies behind the datasets. The FEVER dataset was originally based on the fact verification dataset and consists of factually specific premises that are sufficient for the prediction of the textual entailment. The SNLI dataset, on the other hand, relies more heavily on implicit semantical patterns that the models learn during training, introducing an uncertainty we cannot easily predict.

Our findings are in contrast to the results of~\cite[Section 3, Finding 1]{neg:truong-2023-naysayers}, where it is posited that pre-trained language models perform comparably or worse than a random baseline, and model scaling has almost no effect.
The difference in our findings can be caused by a particular choice of the evaluated models and their level of fine-tuning. A plausible explanation might be that measured accuracy strongly benefits from fine-tuning of prompts.

\paragraph{\textbf{Larger LMs are more sensitive to negation for TE task.}}

Table~\ref{tab:sensitivity} shows the negation sensitivity of each model across individual datasets. Our negation sensitivity metric represents the percentage of datasets in which the model assigned opposite entailment scores to pairs of hypotheses with and without negation. The best results were achieved only by the biggest models, supporting our RQ1.

Similarly to the relative accuracy change, we compute the correlation between model size and the model's sensitivity to negation. As per the computed $p$-values for the Shapiro-Wilk test, we can see that the majority of our negation sensitivity metric across datasets does not have a normal distribution either. Therefore, we are again using the Spearman correlation.

The results indicate a very strong positive correlation of negation sensitivity with the model size on the FEVER datasets, with a rather weak correlation on the SNLI datasets. For an average negation sensitivity for all languages per model, the results indicate a strong positive correlation, with a Spearman correlation coefficient of $0.866$.

Larger models are less likely to classify pair hypotheses with the same textual entailment label, and the impact of negation is more effectively propagated through the model. One outlier in our results is the Mistral (22B) model, which has been outperformed by the Mistral (12B) model on the CES, ENG, and UKR SNLI datasets, as well as on the DEU FEVER dataset.

\begin{table*}
\tabcolsep2.5dd
\begin{tabular}{@{}rlllllllll@{}}
\toprule
\multicolumn{1}{r}{\textbf{Model (size)}} &
  \TT{CES\\FEVER\\MANUAL} &
  \TT{CES\\FEVER} &
  \TT{CES\\SNLI} &
  \TT{ENG\\FEVER} &
  \TT{ENG\\SNLI} &
  \TT{DEU\\FEVER} &
  \TT{DEU\\SNLI} &
  \TT{UKR\\FEVER} &
  \TT{UKR\\SNLI} \\
  \midrule
\M Llama (3B) & 68.04 \% & 68.04 \% & 64.88 \% & 73.54 \% & 81.23 \% & 49.08 \% & 61.85 \% & 65.19 \% & 68.36 \% \\
\M Llama (8B) & 80.5 \% & 80.5 \% & 75.87 \% & 88.62 \% & 86.93 \% & 78.07 \% & 84.46 \% & 80.38 \% & \textbf{79.08 \%} \\
\M Llama (70B) & 87.62 \% & 87.62 \% & \textbf{80.95 \%} & \textbf{94.81 \%} & \textbf{92.29 \%} & \textbf{84.22 \%} & \textbf{89.69 \%} & 84.77 \% & 78.98 \% \\
\M Qwen (1.5B) & 55.92 \% & 55.92 \% & 40.44 \% & 64.77 \% & 62.37 \% & 48.08 \% & 58.58 \% & 54.31 \% & 37.85 \% \\
\M Qwen (7B) & 81.85 \% & 81.85 \% & 71.03 \% & 89.54 \% & 77.35 \% & 70.95 \% & 83.85 \% & 79.46 \% & 72.85 \% \\
\M Qwen (72B) & 87.65 \% & 87.65 \% & 71.83 \% & 92.65 \% & 80.93 \% & 75.56 \% & 88.38 \% & \textbf{85.38 \%} & 70.34 \% \\
\M Mistral (12B) & 80.65 \% & 80.65 \% & 69.4 \% & 87.88 \% & 87.06 \% & 74.8 \% & 83.81 \% & 77.04 \% & 66.19 \% \\
\M Mistral (22B) & 85.23 \% & 85.23 \% & 62.11 \% & 91.46 \% & 66.79 \% & 63.79 \% & 87.08 \% & 77.81 \% & 62.08 \% \\
\M Mistral (123B) & \textbf{87.81 \%} & \textbf{87.81 \%} & 72.1 \% & 91.31 \% & 80.59 \% & 77.3 \% & 89.08 \% & 84.23 \% & 74.09 \% \\
\bottomrule
\end{tabular}
\caption{Evaluation of the negation sensitivity of each model across individual datasets, with the best results bolded.
}
\label{tab:sensitivity}
\end{table*}

\paragraph{\textbf{Language plays a role in the models' accuracy.}} 
Figure~\ref{fig:performancechange} shows the accuracy change on different models with regard to the language of the dataset. 
As the language is a categorical variable, instead of correlation, we compute and compare the variance of the relative accuracy change across the four evaluated languages (CES, ENG, DEU, UKR), excluding the Czech manually translated dataset. We aggregate the values shown in Figure~\ref{fig:fullresults} per language and use the Friedman test as our analysis method. Our null hypothesis is that all languages have the same average of relative accuracy change, and the alternate hypothesis is that the averages are not the same.

\begin{table}[ht]
\tabcolsep2dd
\begin{tabular}{@{}l@{}rrr}
\toprule
    & \TT{ENG} & \TT{DEU} & \TT{UKR}\\
\midrule
CES & 0.030    & 0.0273   & 0.0390\\
ENG &          & 0.4961   & 0.4961\\ 
DEU &          &          & 0.0195\\
\bottomrule
\end{tabular}~~\begin{tabular}{rrr@{}}
\toprule
     \TT{ENG} & \TT{DEU} & \TT{UKR} \\
\midrule
 0.039 & 0.6523 & 0.6523\\
       & 0.2005 & 0.0977\\ 
       &        & 0.2031\\
\bottomrule
\end{tabular}
\caption{$p$-values of Wilcoxon Signed-Test for language pairs on the SNLI (left) and the FEVER (right) datasets}
\label{tab:wilcoxon}
\vspace*{-.5\baselineskip}
\end{table}

Analysis has shown that for our degrees of freedom $d_{\mathit{fn}} = 3$, $d_{\mathit{df}} = 33$, the results are $F$-statistic = $13.000$ and $p$-value = $0.0046$.

As the $p$-value is less than our alpha level, we can reject our null hypothesis. Thus, we conclude that there is a significant difference in the relative accuracy change across different languages.

Delving further and applying the Wilcoxon Signed-Rank Test on the results of language pairs per specific dataset, we find that there is a statistically significant difference on the Czech SNLI dataset and all of the other languages, with the models performing worse on the Czech dataset. On the FEVER dataset, the models performed the worst on the Czech dataset as well, but the difference is statistically significant only in comparison to the English dataset.

Our results partially conform to our presuppositions in RQ2 that the models will perform worse in languages with higher word-order flexibility.

\section{Conclusions and future work}
\vspace*{-.5\baselineskip}
\label{sec:conclusions}
This work presents a novel dataset enabling the evaluation of language models' robustness in handling negations in four languages.
We have extended existing FEVER~\cite{nli:thorne-etal-2018-fever} and SNLI~\cite{nlp:bowman-etal-2015-snli} datasets to include paired hypotheses that are logically opposite to each other by negating the primary verb in the original hypothesis. We have evaluated the models' accuracy separately on \textbf{hypotheses with and without negation} and computed the accuracy drop that occurred due to the addition of negation in the hypothesis. New datasets NoFEVER-ML and NoSNLI-ML are available in the anonymized dataset~\cite{data:datasets}.

We find that model size plays a key role in handling negation correctly, countering the previous work of \citet{neg:truong-2023-naysayers}.
The results on differing language datasets only partially match the initial expectations, with models performing worse on the Czech configuration. This may be due to the relative simplicity of hypotheses, which do not lend to the inclusion of more complex negation structures, such as the negation token being farther away from the verb, as may be the case in German.\looseness=-1

In regard to the differing accuracy from one dataset to another, we hypothesize that the accuracy is also heavily dependent on the specificity of information in the premise; the more common-sense SNLI dataset performed worse than the more fact-explicit FEVER dataset.

Our work provides new resources for addressing the deficiency of language models in handling negations.
Our methodology and dataset can aid in finding the root cause of the negation problem by computing the accuracy on subsets of our new datasets in 4~languages, e.g., a subset of non-projective expressions of negation, a subset of different ways of expressing syntactic or semantic negation~\cite{neg:kalouli-etal-2022-negation}, or to evaluate sub-clausal negation based on a dataset of \citet{nli:truong-etal-2022-sub-clausal-negation}. 
We will use the lens of LLM-microscope~\cite{llm:razzhigaev-2025-llm-microscope-uncoveringhiddenrole} to look at the tokens that play decisive r\^ole.

\iffalse\else\clearpage\fi

\section*{Limitations}
The details of model preparation (training data, hyperparameters) limit the full understanding of the results we report, as we use pretrained models. 
Additionally, this paper does not delve into the effects of post-training and fine-tuning, and possible exposure to classes of negation examples as contradictory usage of negation.
Also, the results might be biased by the semi-automated translations of datasets to their German, Czech, and Ukrainian versions, as seen in Table~\ref{tab:adata-outlier}.
However, we did a manual check to ensure the quality of a subset of the translations and did not encounter any errors related to the application of negation in context.

\clearpage 
\appendix

\section{Full details of accuracy across evaluated models}
\label{appendix:absoluteacc}

\begin{figure*}
\centerline{\includegraphics[width=\textwidth]{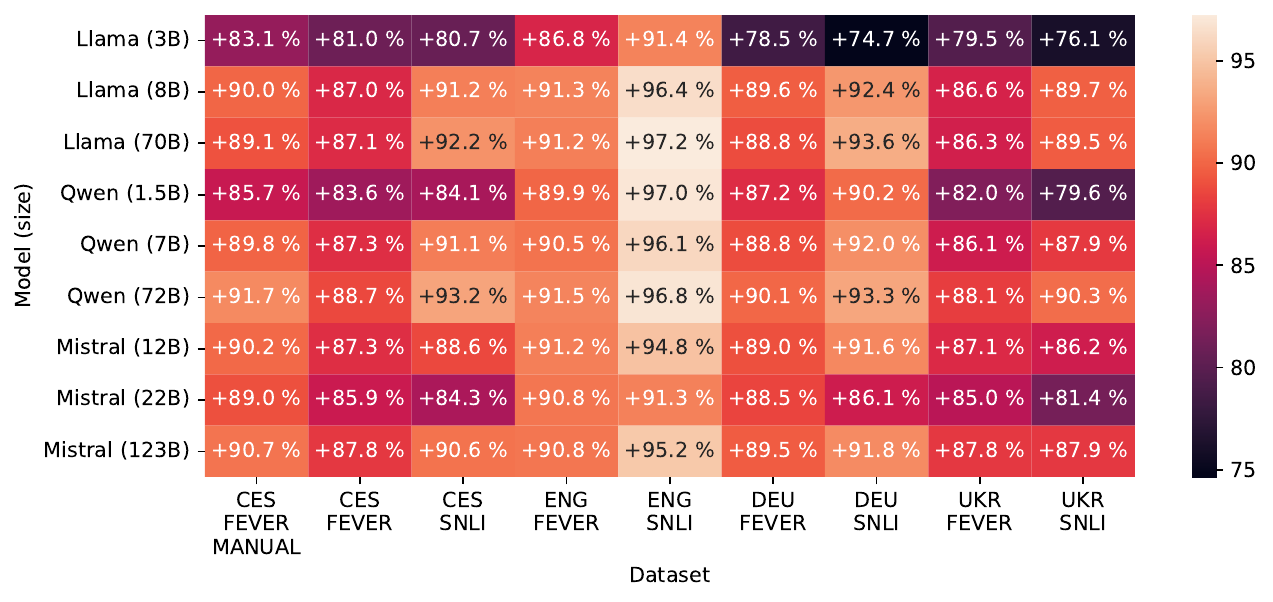}}
\vspace*{-.97\baselineskip}%
\caption{Absolute accuracy for all evaluated models across all evaluated languages and datasets for textual entailment of hypotheses without negation.}
\label{fig:absnoneg}
\bigskip
\centerline{\includegraphics[width=\textwidth]{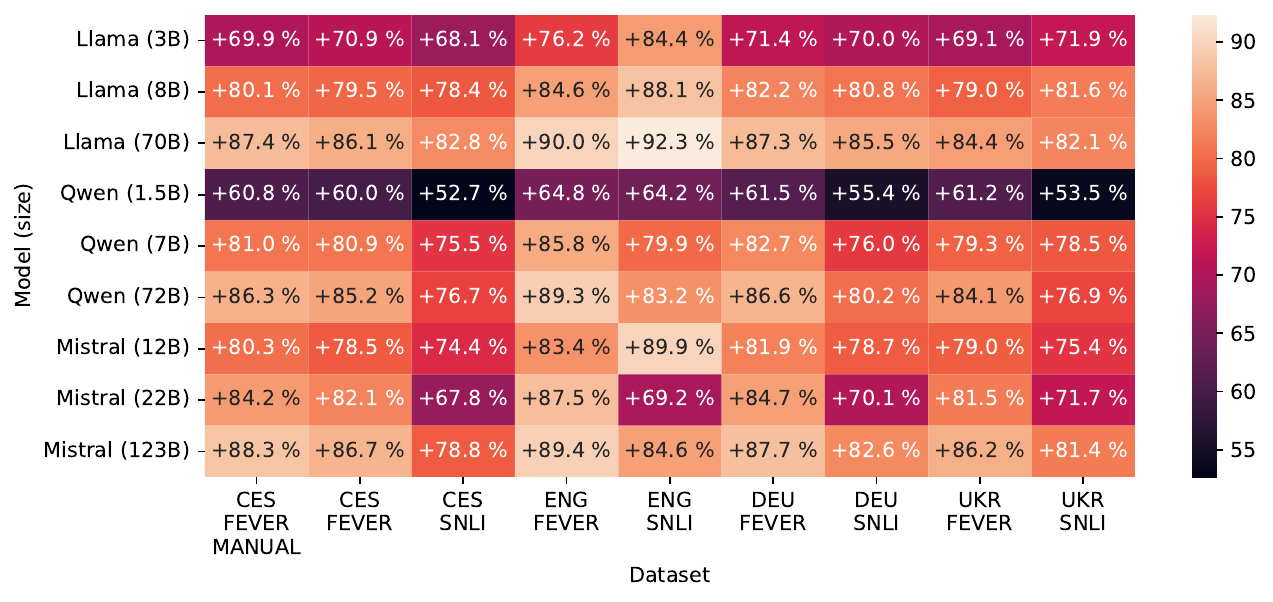}}
\vspace*{-.97\baselineskip}%
\caption{Absolute accuracy for all evaluated models across all evaluated languages and datasets for textual entailment of hypotheses with negation.}
\label{fig:absneg}
\end{figure*}

We include figures~\ref{fig:absnoneg} and \ref{fig:absneg} as the full tables of accuracies of models for hypotheses with and without negation on the task of textual entailment.
In both tables, we can see that almost all models achieve the highest accuracy on the English datasets, on both hypotheses, with and without negation.  

However, the accuracy of the models from the Qwen family shown in Figure~\ref{fig:fullresults} gave us a pause, particularly the fact that overperformance is present only for one specific dataset.
The English SNLI dataset with hypotheses without negation is very similar to the original SNLI dataset, as less than 5~percent of hypotheses in the original test dataset contain negation.
As the original SNLI is a very popular benchmark for the evaluation of NLI and textual entailment tasks, and the Qwen overperformance did not translate across languages on the same dataset, nor did it translate into an increase of accuracy on hypotheses with negation, we have to wonder whether the training data of the Qwen model family inadvertently included the SNLI evaluation data.

\begin{table*}[t]
\tabcolsep3dd
\begin{tabularx}{\textwidth}{@{}Xlllllllll@{}}
\toprule
\multicolumn{1}{r}{\textbf{}} &
  \TT{CES\\FEVER\\MANUAL} &
  \TT{CES\\FEVER} &
  \TT{CES\\SNLI} &
  \TT{ENG\\FEVER} &
  \TT{ENG\\SNLI} &
  \TT{DEU\\FEVER} &
  \TT{DEU\\SNLI} &
  \TT{UKR\\SNLI} &
  \TT{UKR\\FEVER} \\
\midrule
\M Shapiro-Wilk test $p$-value & 0.279 & 0.004 & 0.008 & 0.134 & 0.001 & 0.016 & 0.046 & 0.003 & 0.016 \\
\M Spearman correlation coefficient & 0.867 & 0.883 & 0.400 & 0.883 & 0.250 & 0.900 & 0.300 & 0.883 & 0.150
\end{tabularx}
\caption{Values used during the computation of correlation between the model size and the relative accuracy change across all evaluated datasets.}
\label{tab:statSize}
\end{table*}

\section{Prompting template}
\label{appendix:prompting}
All the models we evaluated support a chat-like interface with a system prompt, a user prompt, and a model response.

As a system prompt, we used the following template:

\begin{quote}
\texttt{%
You are a fact checker for queries in the \textbf{<English / Czech / German / Ukrainian>} language. You will be given a premise, which you know is factually correct, and a hypothesis.  You will return the truth value of the hypothesis, based on the premise. Return True if the hypothesis is correct and False if the hypothesis is incorrect.
}
\end{quote}

The user prompt follows a simple format:

\begin{quote}
\texttt{%
Premise: \textbf{<premise>} \newline
Hypothesis: \textbf{<hypothesis>}
}
\end{quote}

The model decides by finishing a prepared response:

\begin{quote}
\texttt{%
The answer is: \textbf{<True / False>}
}
\end{quote}

\section{Statistical tests}
\label{appendix:stats}
In tables~\ref{tab:statSize} and \ref{tab:statSensitivity} we provide the values used during statistical computations explained in Section~\ref{sec:results}.

\begin{table*}[t]
\tabcolsep2.4dd
\begin{tabularx}{\textwidth}{@{}Xlllllllll@{}}
\toprule
\multicolumn{1}{r}{\textbf{}} &
  \TT{CES\\FEVER\\MANUAL} &
  \TT{CES\\FEVER} &
  \TT{CES\\SNLI} &
  \TT{ENG\\FEVER} &
  \TT{ENG\\SNLI} &
  \TT{DEU\\FEVER} &
  \TT{DEU\\SNLI} &
  \TT{UKR\\SNLI} &
  \TT{UKR\\FEVER} \\
\midrule
\M Shapiro-Wilk $p$-value & 0.013 & 0.013 & 0.052 & 0.008 & 0.405 & 0.101 & 0.002 & 0.027 & 0.022 \\
\M Spearman correlation w/ model size & 0.950 & 0.950 & 0.583 & 0.817 & 0.267 & 0.683 & 0.900 & 0.817 & 0.400
\end{tabularx}
\caption{Values used during the computation of correlation between the model size and the negation sensitivity across all evaluated datasets.}
\label{tab:statSensitivity}
\end{table*}

\section{Sentence complexity}
\label{appendix:complexity}

As mentioned in Section~\ref{sec:results}, we posit that the larger models can more accurately capture nuanced language patterns, enabling them to correctly handle negations in more complex sentences that elude the smaller models. From each model family, we take the smallest and the largest models and compare the average sentence complexity of the hypotheses that the model is able to correctly classify. However, we hypothesize that many hypotheses correctly classified by the smaller model are also correctly classified by the larger model. Therefore, for the larger model, we compute the average sentence complexity of the additional correctly classified hypotheses, i.e., the subset for the larger model constitutes of hypotheses the larger model correctly classified which are not present in the smaller model's subset.\looseness=-1

We evaluate the sentence complexity using the following metrics:

\paragraph{\textbf{Hypothesis depth.}} Depth of the hypothesis dependency tree. Sentence complexity increases with the increase in the depth of the dependency tree.

Tables~\ref{tab:avgDepthPos} and~\ref{tab:avgDepthNeg} show the increase of average hypothesis depth between the smallest and the largest model per the language model family. Due to the relative simplicity of the hypotheses (short length as described in Table~\ref{tab:preparedNLIdata}, average of one verb per hypothesis), the increase of depth is not substantial, being less than one level across all datasets for all language model families.

\paragraph{\textbf{Dissimilarity of hypothesis to the premise.}} Number of content words present only in the hypothesis and not the premise, divided by the total number of content words in the hypothesis. Sentence complexity increases with the increase of dissimilarity. We compute the similarity on the lemmatized forms of the words, with the most common stop words per language filtered out. High dissimilarity may be caused by the usage of synonyms, hypernyms, hyponyms, or other linguistic features that require a higher understanding of syntactic patterns in the hypothesis. 

Tables~\ref{tab:avgSimPos} and~\ref{tab:avgSimNeg} show the increase in the average dissimilarity of the premise and its hypothesis between the smallest and the largest model per the language model family. Here we can see a marked increase in dissimilarity for hypotheses with negation, especially within the Qwen family models, showcasing the largest model's increased ability to correctly reason and determine entailment on texts that use different formulations to express the same meaning.

\begin{table*}[p]
\tabcolsep7dd
\begin{tabularx}{\textwidth}{@{}Xlllllllll@{}}
\toprule
\multicolumn{1}{r}{\textbf{}} &
  \TT{CES\\FEVER\\MANUAL} &
  \TT{CES\\FEVER} &
  \TT{CES\\SNLI} &
  \TT{ENG\\FEVER} &
  \TT{ENG\\SNLI} &
  \TT{DEU\\FEVER} &
  \TT{DEU\\SNLI} &
  \TT{UKR\\FEVER} &
  \TT{UKR\\SNLI} \\
\midrule
\M Llama Family & 0.027 & 0.215 & \textbf{0.064} & 0.270 & \textbf{0.196} & 0.219 & \textbf{0.056} & \textbf{0.155} & \textbf{0.049} \\
\M Qwen Family & \textbf{0.082} & 0.173 & 0.006 & 0.215 & 0.174 & 0.024 & 0.030 & 0.148 & 0.041 \\
\M Mistral Family & \LM-0.016 & \textbf{0.184} & 0.037 & \textbf{0.271} & 0.137 & \textbf{0.245} & 0.001 & 0.121 & \LM-0.017 \\       
\bottomrule
\end{tabularx}
\caption{Increase of average depth of \textbf{hypotheses without negation} that were correctly predicted only by the largest model of the LM family and the correctly predicted hypotheses by the smallest model of the LM family.}
\label{tab:avgDepthPos}
\end{table*}

\begin{table*}[t]
\tabcolsep7dd
\begin{tabularx}{\textwidth}{@{}Xlllllllll@{}}
\toprule
\multicolumn{1}{r}{\textbf{}} &
  \TT{CES\\FEVER\\MANUAL} &
  \TT{CES\\FEVER} &
  \TT{CES\\SNLI} &
  \TT{ENG\\FEVER} &
  \TT{ENG\\SNLI} &
  \TT{DEU\\FEVER} &
  \TT{DEU\\SNLI} &
  \TT{UKR\\FEVER} &
  \TT{UKR\\SNLI} \\
\midrule
\M Llama Family & 0.019 & \textbf{0.076} & \textbf{0.090} & 0.089 & \textbf{0.145} & 0.041 & \textbf{0.030} & 0.150 & \textbf{0.067} \\
\M Qwen Family & \textbf{0.038} & \LM-0.09 & 0.027 & \LM-0.016 & 0.065 & \LM-0.053 & \LM-0.084 & \LM-0.006 & \LM-0.080 \\
\M Mistral Family & \LM-0.016 & 0.031 & 0.064 & \textbf{0.125} & 0.101 & \textbf{0.103} & \LM-0.012 & \textbf{0.151} & \LM-0.017 \\  
\bottomrule
\end{tabularx}
\caption{Increase of average depth of \textbf{hypotheses with negation} that were correctly predicted only by the largest model of the LM family and the correctly predicted hypotheses by the smallest model of the LM family.}
\label{tab:avgDepthNeg}
\end{table*}

\begin{table*}[t]
\tabcolsep4dd
\works
\begin{tabularx}{\textwidth}{@{}Xlllllllll@{}}
\toprule
\multicolumn{1}{r}{\textbf{}} &
  \TT{CES\\FEVER\\MANUAL} &
  \TT{CES\\FEVER} &
  \TT{CES\\SNLI} &
  \TT{ENG\\FEVER} &
  \TT{ENG\\SNLI} &
  \TT{DEU\\FEVER} &
  \TT{DEU\\SNLI} &
  \TT{UKR\\FEVER} &
  \TT{UKR\\SNLI} \\
\midrule
\M Llama Family & \textbf{2.52 \%} & \LM-0.12 \% & \LM-4.31 \% & \LM-5.77 \% & \LM-15.30 \%_& \LM-4.73 \% & \LM-16.01 \%_& \LM-1.20 \% & \LM-6.83 \% \\
\M Qwen Family & \LM-0.47 \% & 4.19 \% & \textbf{2.55 \%} & \textbf{7.88 \%} & _\LM-5.00 \% & 0.90 \% & _3.58 \% & \textbf{4.66 \%} & \textbf{1.44 \%} \\
\M Mistral Family & 0.89 \% & \textbf{4.49 \%} & 0.42 \% & 6.94 \% & \textbf{_\LM-1.63 \%} & \textbf{1.02 \%} & \textbf{_6.26 \%} & 2.49 \%_& \LM-1.92 \% \\
\bottomrule
\end{tabularx}
\caption{Increase of average sentence dissimilarity of premises and the evaluated \textbf{hypotheses without negation} that were correctly predicted only by the largest model of the LM family and the correctly predicted hypotheses by the smallest model of the LM family.}
\label{tab:avgSimPos}
\end{table*}

\begin{table*}[t]
\works\tabcolsep3dd
\begin{tabularx}{\textwidth}{@{}Xlllllllll@{}}
\toprule
\multicolumn{1}{r}{\textbf{}} &
  \TT{CES\\FEVER\\MANUAL} &
  \TT{CES\\FEVER} &
  \TT{CES\\SNLI} &
  \TT{ENG\\FEVER} &
  \TT{ENG\\SNLI} &
  \TT{DEU\\FEVER} &
  \TT{DEU\\SNLI} &
  \TT{UKR\\FEVER} &
  \TT{UKR\\SNLI} \\
\midrule
\M Llama Family & \textbf{2.00 \%} & _3.97 \% & _\LM-3.02 \% & _2.36 \% & _0.62 \% & _5.52 \% & 21.12 \% & 5.56 \% & _\LM-1.52 \% \\
\M Qwen Family & 0.24 \% & \textbf{12.49 \%} & \textbf{19.41 \%} & \textbf{12.50 \%} & \textbf{21.14 \%} & \textbf{10.56 \%} & \textbf{25.16 \%} & \textbf{9.44 \%} & \textbf{15.28 \%} \\
\M Mistral Family & 1.59 \% & _4.06 \% & 11.77\% & _5.22 \% & _6.11 \% & _5.30 \% & 11.96 \% & 4.94 \% & 10.36 \% \\
\bottomrule
\end{tabularx}
\caption{Increase of average sentence dissimilarity of premises and the evaluated \textbf{hypotheses with negation} that were correctly predicted only by the largest model of the LM family and the correctly predicted hypotheses by the smallest model of the LM family.}
\label{tab:avgSimNeg}
\end{table*}


\begin{thebibliography}{36}
\providecommand{\natexlab}[1]{#1}

\bibitem[{Alhamoud et~al.(2025)Alhamoud, Alshammari, Tian, Li, Torr, Kim, and Ghassemi}]{neg:alhamoud2025visionlanguagemodelsunderstandnegation}
Kumail Alhamoud, Shaden Alshammari, Yonglong Tian, Guohao Li, Philip Torr, Yoon Kim, and Marzyeh Ghassemi. 2025.
\newblock \href {https://arxiv.org/abs/2501.09425} {{Vision-Language Models Do Not Understand Negation}}.
\newblock \emph{Preprint}, arXiv:2501.09425.

\bibitem[{{Anonymous}(2025)}]{data:datasets}
{Anonymous}. 2025.
\newblock \href {https://doi.org/10.5281/zenodo.15455465} {{Datasets for submission}}.
\newblock Zenodo.

\bibitem[{Ansch{\"u}tz et~al.(2023{\natexlab{a}})Ansch{\"u}tz, Miguel~Lozano, and Groh}]{neg:anschutz-etal-2023-correct}
Miriam Ansch{\"u}tz, Diego Miguel~Lozano, and Georg Groh. 2023{\natexlab{a}}.
\newblock \href {https://doi.org/10.18653/v1/2023.inlg-main.12} {{This is not correct! Negation-aware Evaluation of Language Generation Systems}}.
\newblock In \emph{Proceedings of the 16th International Natural Language Generation Conference}, pages 163--175, Prague, Czechia. ACL.

\bibitem[{Ansch{\"u}tz et~al.(2023{\natexlab{b}})Ansch{\"u}tz, Miguel~Lozano, and Groh}]{nli:anschutz-etal-2023-correct}
Miriam Ansch{\"u}tz, Diego Miguel~Lozano, and Georg Groh. 2023{\natexlab{b}}.
\newblock \href {https://doi.org/10.18653/v1/2023.inlg-main.12} {{This is not correct! Negation-aware Evaluation of Language Generation Systems}}.
\newblock In \emph{Proceedings of the 16th International Natural Language Generation Conference}, pages 163--175, Prague, Czechia. ACL.

\bibitem[{Bowman et~al.(2015)Bowman, Angeli, Potts, and Manning}]{nlp:bowman-etal-2015-snli}
Samuel~R. Bowman, Gabor Angeli, Christopher Potts, and Christopher~D. Manning. 2015.
\newblock \href {https://doi.org/10.18653/v1/D15-1075} {A large annotated corpus for learning natural language inference}.
\newblock In \emph{Proceedings of the 2015 Conference on Empirical Methods in Natural Language Processing}, pages 632--642, Lisbon, Portugal. ACL.

\bibitem[{Castricato et~al.(2024)Castricato, Lile, Anand, Schoelkopf, Verma, and Biderman}]{nlp:castricato-2024-suppressing-pink-elephants-direct}
Louis Castricato, Nathan Lile, Suraj Anand, Hailey Schoelkopf, Siddharth Verma, and Stella Biderman. 2024.
\newblock \href {https://arxiv.org/abs/2402.07896} {{Suppressing Pink Elephants with Direct Principle Feedback}}.
\newblock \emph{Preprint}, arXiv:2402.07896.

\bibitem[{Dziri et~al.(2023)Dziri, Lu, Sclar, Li, Jiang, Lin, Welleck, West, Bhagavatula, Bras, Hwang, Sanyal, Ren, Ettinger, Harchaoui, and Choi}]{nlp:dziri-2023-faith-fate}
Nouha Dziri, Ximing Lu, Melanie Sclar, Xiang~Lorraine Li, Liwei Jiang, Bill~Yuchen Lin, Sean Welleck, Peter West, Chandra Bhagavatula, Ronan~Le Bras, Jena~D. Hwang, Soumya Sanyal, Xiang Ren, Allyson Ettinger, Zaid Harchaoui, and Yejin Choi. 2023.
\newblock \href {https://openreview.net/forum?id=Fkckkr3ya8} {{Faith and Fate: Limits of Transformers on Compositionality}}.
\newblock In \emph{Thirty-seventh Conference on Neural Information Processing Systems}.

\bibitem[{Ettinger(2020)}]{neg:ettinger-2020-what-bert-is-not}
Allyson Ettinger. 2020.
\newblock \href {https://doi.org/10.1162/tacl\_a\_00298} {{What BERT Is Not: Lessons from a New Suite of Psycholinguistic Diagnostics for Language Models}}.
\newblock \emph{Transactions of the Association for Computational Linguistics}, 8:34--48.

\bibitem[{Garc{\'i}a-Ferrero et~al.(2023)Garc{\'i}a-Ferrero, Altuna, Alvez, Gonzalez-Dios, and Rigau}]{neg:garcia-ferrero-etal-2023-dataset}
Iker Garc{\'i}a-Ferrero, Bego{\~n}a Altuna, Javier Alvez, Itziar Gonzalez-Dios, and German Rigau. 2023.
\newblock \href {https://doi.org/10.18653/v1/2023.emnlp-main.531} {{This is not a Dataset: A Large Negation Benchmark to Challenge Large Language Models}}.
\newblock In \emph{Proceedings of the 2023 Conference on Empirical Methods in Natural Language Processing}, pages 8596--8615, Singapore. ACL.

\bibitem[{Gat(2024)}]{lm_format_enforcer}
Noam Gat. 2024.
\newblock {LM-Format-Enforcer: A text formatter for language models}.
\newblock \url{https://github.com/noamgat/lm-format-enforcer}.
\newblock Accessed: 2024-12-16.

\bibitem[{Gururangan et~al.(2018)Gururangan, Swayamdipta, Levy, Schwartz, Bowman, and Smith}]{neg:gururangan-etal-2018-annotation}
Suchin Gururangan, Swabha Swayamdipta, Omer Levy, Roy Schwartz, Samuel Bowman, and Noah~A. Smith. 2018.
\newblock \href {https://doi.org/10.18653/v1/N18-2017} {{Annotation Artifacts in Natural Language Inference Data}}.
\newblock In \emph{Proceedings of the 2018 Conference of the North {A}merican Chapter of the Association for Computational Linguistics: Human Language Technologies, Volume 2 (Short Papers)}, pages 107--112, New Orleans, Louisiana. ACL.

\bibitem[{Hartmann et~al.(2021)Hartmann, de~Lhoneux, Hershcovich, Kementchedjhieva, Nielsen, Qiu, and S{\o}gaard}]{neg:hartmann-etal-2021-multilingual}
Mareike Hartmann, Miryam de~Lhoneux, Daniel Hershcovich, Yova Kementchedjhieva, Lukas Nielsen, Chen Qiu, and Anders S{\o}gaard. 2021.
\newblock \href {https://doi.org/10.18653/v1/2021.conll-1.19} {{A Multilingual Benchmark for Probing Negation-Awareness with Minimal Pairs}}.
\newblock In \emph{Proceedings of the 25th Conference on Computational Natural Language Learning}, pages 244--257, Online. ACL.

\bibitem[{Helwe et~al.(2022)Helwe, Coumes, Clavel, and Suchanek}]{helwe-etal-2022-tina}
Chadi Helwe, Simon Coumes, Chlo{\'e} Clavel, and Fabian Suchanek. 2022.
\newblock \href {https://doi.org/10.18653/v1/2022.findings-emnlp.301} {{{TINA}: Textual Inference with Negation Augmentation}}.
\newblock In \emph{Findings of the Association for Computational Linguistics: EMNLP 2022}, pages 4086--4099, Abu Dhabi, United Arab Emirates. ACL.

\bibitem[{Hermann et~al.(2013)Hermann, Grefenstette, and Blunsom}]{nlp:hermann-etal-2013-bad}
Karl~Moritz Hermann, Edward Grefenstette, and Phil Blunsom. 2013.
\newblock \href {https://aclanthology.org/W13-3209} {{``N}ot not bad{''} is not {``}bad{''}: {A} distributional account of negation}.
\newblock In \emph{Proceedings of the Workshop on Continuous Vector Space Models and their Compositionality}, pages 74--82, Sofia, Bulgaria. ACL.

\bibitem[{Kalouli et~al.(2022)Kalouli, Sevastjanova, Beck, and Romero}]{neg:kalouli-etal-2022-negation}
Aikaterini-Lida Kalouli, Rita Sevastjanova, Christin Beck, and Maribel Romero. 2022.
\newblock \href {https://aclanthology.org/2022.coling-1.272/} {{Negation, Coordination, and Quantifiers in Contextualized Language Models}}.
\newblock In \emph{Proceedings of the 29th International Conference on Computational Linguistics}, pages 3074--3085, Gyeongju, Republic of Korea. ACL.

\bibitem[{Kassner and Sch{\"u}tze(2020)}]{neg:kassner-schutze-2020-negated-probes}
Nora Kassner and Hinrich Sch{\"u}tze. 2020.
\newblock \href {https://doi.org/10.18653/v1/2020.acl-main.698} {Negated and misprimed probes for pretrained language models: Birds can talk, but cannot fly}.
\newblock In \emph{Proceedings of the 58th Annual Meeting of the ACL}, pages 7811--7818, Online. ACL.

\bibitem[{{Llama Team}(2024{\natexlab{a}})}]{data:llama3-4negation}
{Llama Team}. 2024{\natexlab{a}}.
\newblock \href {https://huggingface.co/meta-llama/} {{Llama AI Models on Hugging Face}}.
\newblock Hugging Face Model Repository.
\newblock Accessed [2024-12-15]. Models: \href{https://huggingface.co/meta-llama/Llama-3.2-3B-Instruct}{3B}, \href{https://huggingface.co/meta-llama/Llama-3.1-8B-Instruct}{8B}, \href{https://huggingface.co/meta-llama/Llama-3.3-70B-Instruct}{70B}.

\bibitem[{{Llama Team}(2024{\natexlab{b}})}]{nlp:Dubey-etal-2024-llama3-herd-of-models}
{Llama Team}. 2024{\natexlab{b}}.
\newblock \href {https://api.semanticscholar.org/CorpusID:271571434} {{The Llama 3 Herd of Models}}.
\newblock \emph{ArXiv}, abs/2407.21783.

\bibitem[{{Mistral AI}(2024)}]{data:mistral-hf-4negation}
{Mistral AI}. 2024.
\newblock \href {https://huggingface.co/mistralai} {{Mistral AI Models on Hugging Face}}.
\newblock Hugging Face Model Repository.
\newblock Accessed [2024-12-15]. Models: \href{https://huggingface.co/mistralai/Mistral-Nemo-Instruct-2407}{12B Nemo}, \href{https://huggingface.co/mistralai/Mistral-Small-Instruct-2409}{22B Small}, \href{https://huggingface.co/mistralai/Mistral-Large-Instruct-2411}{123B Large}.

\bibitem[{Mondorf and Plank(2024)}]{neg:mondorf-plank-2024-liar}
Philipp Mondorf and Barbara Plank. 2024.
\newblock \href {https://doi.org/10.18653/v1/2024.emnlp-main.404} {{Liar, Liar, Logical Mire: A Benchmark for Suppositional Reasoning in Large Language Models}}.
\newblock In \emph{Proceedings of the 2024 Conference on Empirical Methods in Natural Language Processing}, pages 7114--7137, Miami, Florida, USA. ACL.

\bibitem[{Nadeem et~al.(2024)Nadeem, Sohail, Cambria, Schuller, and Hussain}]{neg:nadeem2024negationblindnesslargelanguage}
Mohammad Nadeem, Shahab~Saquib Sohail, Erik Cambria, Björn~W. Schuller, and Amir Hussain. 2024.
\newblock \href {https://arxiv.org/abs/2409.00105} {{Negation Blindness in Large Language Models: Unveiling the NO Syndrome in Image Generation}}.
\newblock \emph{Preprint}, arXiv:2409.00105.

\bibitem[{Putra et~al.(2024)Putra, Siahaan, and Saikhu}]{nli:Putra-2024-textual-entailment-overview}
I~Made~Suwija Putra, Daniel Siahaan, and Ahmad Saikhu. 2024.
\newblock \href {https://doi.org/10.1016/j.icte.2023.08.012} {Recognizing textual entailment: A review of resources, approaches, applications, and challenges}.
\newblock \emph{ICT Express}, 10(1):132--155.

\bibitem[{{Qwen Team}(2024)}]{data:qwen2.5}
{Qwen Team}. 2024.
\newblock \href {https://qwenlm.github.io/blog/qwen2.5/} {Qwen2.5: A party of foundation models}.

\bibitem[{Razzhigaev et~al.(2025)Razzhigaev, Mikhalchuk, Rahmatullaev, Goncharova, Druzhinina, Oseledets, and Kuznetsov}]{llm:razzhigaev-2025-llm-microscope-uncoveringhiddenrole}
Anton Razzhigaev, Matvey Mikhalchuk, Temurbek Rahmatullaev, Elizaveta Goncharova, Polina Druzhinina, Ivan Oseledets, and Andrey Kuznetsov. 2025.
\newblock \href {https://aclanthology.org/2025.findings-naacl.432/} {{LLM-Microscope: Uncovering the Hidden Role of Punctuation in Context Memory of Transformers}}.
\newblock In \emph{Findings of the ACL: NAACL 2025}, pages 7757--7764, Albuquerque, New Mexico. ACL.

\bibitem[{Rezaei and Blanco(2024)}]{rezaei-blanco-2024-paraphrasing}
MohammadHossein Rezaei and Eduardo Blanco. 2024.
\newblock \href {https://doi.org/10.18653/v1/2024.acl-short.55} {Paraphrasing in affirmative terms improves negation understanding}.
\newblock In \emph{Proceedings of the 62nd Annual Meeting of the Association for Computational Linguistics (Volume 2: Short Papers)}, pages 602--615, Bangkok, Thailand. ACL.

\bibitem[{Singh et~al.(2024)Singh, Shrivastava, Vatsa, Singh, and Bharati}]{neg:singh2024learnnosayyes}
Jaisidh Singh, Ishaan Shrivastava, Mayank Vatsa, Richa Singh, and Aparna Bharati. 2024.
\newblock \href {https://doi.org/10.48550/arXiv.2403.20312} {{Learn "No" to Say "Yes" Better: Improving Vision-Language Models via Negations}}.
\newblock \emph{Preprint}, arXiv:2403.20312.

\bibitem[{Spiers(2002)}]{psy:spiers-2002-pink-elephant-paradox}
Jude~A. Spiers. 2002.
\newblock \href {https://doi.org/10.1177/160940690200100405} {{The Pink Elephant Paradox (or, Avoiding the Misattribution of Data)}}.
\newblock \emph{International Journal of Qualitative Methods}, 1(4):36--44.

\bibitem[{Thorne et~al.(2018)Thorne, Vlachos, Christodoulopoulos, and Mittal}]{nli:thorne-etal-2018-fever}
James Thorne, Andreas Vlachos, Christos Christodoulopoulos, and Arpit Mittal. 2018.
\newblock \href {https://doi.org/10.18653/v1/N18-1074} {{{FEVER}: a Large-scale Dataset for Fact Extraction and {VER}ification}}.
\newblock In \emph{Proceedings of the 2018 Conference of the North {A}merican Chapter of the Association for Computational Linguistics: Human Language Technologies, Volume 1 (Long Papers)}, pages 809--819, New Orleans, Louisiana. ACL.

\bibitem[{Truong et~al.(2023)Truong, Baldwin, Verspoor, and Cohn}]{neg:truong-2023-naysayers}
Thinh~Hung Truong, Timothy Baldwin, Karin Verspoor, and Trevor Cohn. 2023.
\newblock \href {https://doi.org/10.18653/v1/2023.starsem-1.10} {Language models are not naysayers: An analysis of language models on negation benchmarks}.
\newblock In \emph{Proc. of the 12th Joint Conference on Lexical and Computational Semantics (*SEM 2023)}, pages 101--114, Toronto, Canada. ACL.

\bibitem[{Truong et~al.(2022)Truong, Otmakhova, Baldwin, Cohn, Lau, and Verspoor}]{nli:truong-etal-2022-sub-clausal-negation}
Thinh~Hung Truong, Yulia Otmakhova, Timothy Baldwin, Trevor Cohn, Jey~Han Lau, and Karin Verspoor. 2022.
\newblock \href {https://doi.org/10.18653/v1/2022.aacl-main.65} {{Not another Negation Benchmark: The {N}a{N}-{NLI} Test Suite for Sub-clausal Negation}}.
\newblock In \emph{Proceedings of the 2nd Conference of the Asia-Pacific Chapter of the Association for Computational Linguistics and the 12th International Joint Conference on Natural Language Processing (Volume 1: Long Papers)}, pages 883--894, Online only. ACL.

\bibitem[{Varshney et~al.(2025)Varshney, Raj, Mishra, Chatterjee, Saeidi, Sarkar, and Baral}]{neg:varshney-etal-2025-investigating}
Neeraj Varshney, Satyam Raj, Venkatesh Mishra, Agneet Chatterjee, Amir Saeidi, Ritika Sarkar, and Chitta Baral. 2025.
\newblock \href {https://aclanthology.org/2025.trustnlp-main.37/} {{Investigating and Addressing Hallucinations of {LLM}s in Tasks Involving Negation}}.
\newblock In \emph{Proceedings of the 5th Workshop on Trustworthy NLP (TrustNLP 2025)}, pages 580--598, Albuquerque, New Mexico. Association for Computational Linguistics.

\bibitem[{Vaswani et~al.(2017)Vaswani, Shazeer, Parmar, Uszkoreit, Jones, Gomez, Kaiser, and Polosukhin}]{llm:vaswani2017attention}
Ashish Vaswani, Noam Shazeer, Niki Parmar, Jakob Uszkoreit, Llion Jones, Aidan~N Gomez, {\L}ukasz Kaiser, and Illia Polosukhin. 2017.
\newblock \href {http://arxiv.org/abs/1706.03762v7} {Attention is all you need}.
\newblock In \emph{Advances in neural information processing systems}, pages 5998--6008.

\bibitem[{Vrabcová and Sojka(2024)}]{neg:vrabcova-sojka-2024-negation-czech}
Tereza Vrabcová and Petr Sojka. 2024.
\newblock \href {https://nlp.fi.muni.cz/raslan/2024/paper11.pdf} {{Negation Disrupts Compositionality in~Language~Models: The Czech Usecase}}.
\newblock In \emph{Proceedings of the 18th Workshop on Recent Advances in Slavonic Natural Language Processing, RASLAN 2024, Kouty nad Desnou, Czech Republic, December 6--8, 2024}, pages 17--24, Brno. Tribun EU.

\bibitem[{Willard and Louf(2023)}]{nlp:willard2023efficient}
Brandon~T Willard and R{\'e}mi Louf. 2023.
\newblock \href {https://doi.org/10.48550/arXiv.2307.09702} {{Efficient Guided Generation for Large Language Models}}.
\newblock \emph{arXiv preprint arXiv:2307.09702}.

\bibitem[{Yang et~al.(2024)Yang, Yang, Hui, Zheng, Yu, Zhou, Li, Li, Liu, Huang, Dong, Wei, Lin, Tang, Wang, Yang, Tu, Zhang, Ma, Yang, Xu, Zhou, Bai, He, Lin, Dang, Lu, Chen, Yang, Li, Xue, Ni, Zhang, Wang, Peng, Men, Gao, Lin, Wang, Bai, Tan, Zhu, Li, Liu, Ge, Deng, Zhou, Ren, Zhang, Wei, Ren, Liu, Fan, Yao, Zhang, Wan, Chu, Liu, Cui, Zhang, Guo, and Fan}]{nlp:yang-etal-2024-qwen2-technicalreport}
An~Yang, Baosong Yang, Binyuan Hui, Bo~Zheng, Bowen Yu, Chang Zhou, Chengpeng Li, Chengyuan Li, Dayiheng Liu, Fei Huang, Guanting Dong, Haoran Wei, Huan Lin, Jialong Tang, Jialin Wang, Jian Yang, Jianhong Tu, Jianwei Zhang, Jianxin Ma, Jianxin Yang, Jin Xu, Jingren Zhou, Jinze Bai, Jinzheng He, Junyang Lin, Kai Dang, Keming Lu, Keqin Chen, Kexin Yang, Mei Li, Mingfeng Xue, Na~Ni, Pei Zhang, Peng Wang, Ru~Peng, Rui Men, Ruize Gao, Runji Lin, Shijie Wang, Shuai Bai, Sinan Tan, Tianhang Zhu, Tianhao Li, Tianyu Liu, Wenbin Ge, Xiaodong Deng, Xiaohuan Zhou, Xingzhang Ren, Xinyu Zhang, Xipin Wei, Xuancheng Ren, Xuejing Liu, Yang Fan, Yang Yao, Yichang Zhang, Yu~Wan, Yunfei Chu, Yuqiong Liu, Zeyu Cui, Zhenru Zhang, Zhifang Guo, and Zhihao Fan. 2024.
\newblock \href {https://doi.org/10.48550/arXiv.2407.10671} {{Qwen2 Technical Report}}.
\newblock \emph{Preprint}, arXiv:2407.10671.

\bibitem[{Zhang et~al.(2023)Zhang, Yasunaga, Zhou, HaoChen, Zou, Liang, and Yeung}]{neg:zhang-etal-2023-beyond}
Yuhui Zhang, Michihiro Yasunaga, Zhengping Zhou, Jeff~Z. HaoChen, James Zou, Percy Liang, and Serena Yeung. 2023.
\newblock \href {https://doi.org/10.18653/v1/2023.findings-acl.472} {{Beyond Positive Scaling: How Negation Impacts Scaling Trends of Language Models}}.
\newblock In \emph{Findings of the Association for Computational Linguistics: ACL 2023}, pages 7479--7498, Toronto, Canada. ACL.

\end{thebibliography}
\end{document}